\pdfoutput=1

\documentclass[11pt]{article}

\usepackage[final]{acl}

\usepackage{times}
\usepackage{latexsym}

\usepackage[T1]{fontenc}

\usepackage[utf8]{inputenc}

\usepackage{microtype}

\usepackage{inconsolata}

\usepackage{graphicx}

\usepackage{lipsum}
\usepackage{booktabs}
\usepackage{multirow}
\usepackage{graphicx}
\usepackage{tabularx}
\usepackage{subfig}
\usepackage[export]{adjustbox}
\usepackage{amsmath}
\usepackage{enumitem}
\usepackage{float}
\usepackage{placeins}
\usepackage{afterpage}
\usepackage{titlesec}
\usepackage{tcolorbox}
\usepackage{colortbl}
\usepackage{array, xcolor}
\usepackage{tabularray}

\title{LLM-as-a-tutor in EFL Writing Education: \\Focusing on Evaluation of Student-LLM Interaction}

\author{
    Jieun Han, Haneul Yoo, Junho Myung, Minsun Kim, Hyunseung Lim, Yoonsu Kim, \\
    \textbf{Tak Yeon Lee, Hwajung Hong, Juho Kim, So-Yeon Ahn, Alice Oh} \\
    KAIST, South Korea \\
    \texttt{\{\href{mailto:jieun_han@kaist.ac.kr}{\color{black}{jieun\_han}}, \href{mailto:haneul.yoo@kaist.ac.kr}{\color{black}{haneul.yoo}}, \href{mailto:junho00211@kaist.ac.kr}{\color{black}{junho00211}}, \href{mailto:9909cindy@kaist.ac.kr}{\color{black}{9909cindy}}, 
    \href{mailto:charlie9807@kaist.ac.kr}{\color{black}{charlie9807}},
    \href{mailto:yoonsu16@kaist.ac.kr}{\color{black}{yoonsu16}},}\\
    \texttt{
    \href{mailto:takyeonlee@kaist.ac.kr}{\color{black}{takyeonlee}},
    \href{mailto:hwajung@kaist.ac.kr}{\color{black}{hwajung}},
    \href{mailto:juhokim@kaist.ac.kr}{\color{black}{juhokim}},
    \href{mailto:ahnsoyeon@kaist.ac.kr}{\color{black}{ahnsoyeon}}\}@kaist.ac.kr},
    \texttt{alice.oh@kaist.edu}
}

\begin{document}
\maketitle

\begin{abstract}
In the context of English as a Foreign Language (EFL) writing education, LLM-as-a-tutor can assist students by providing real-time feedback on their essays.
However, challenges arise in assessing LLM-as-a-tutor due to differing standards between educational and general use cases. 
To bridge this gap, we integrate pedagogical principles to assess student-LLM interaction.
First, we explore how LLMs can function as English tutors, providing effective essay feedback tailored to students.
Second, we propose three metrics to evaluate LLM-as-a-tutor specifically designed for EFL writing education, emphasizing pedagogical aspects. 
In this process, EFL experts evaluate the feedback from LLM-as-a-tutor regarding \textit{quality} and \textit{characteristics}. On the other hand, EFL learners assess their \textit{learning outcomes} from interaction with LLM-as-a-tutor.
This approach lays the groundwork for developing LLMs-as-a-tutor tailored to the needs of EFL learners, advancing the effectiveness of writing education in this context.
\end{abstract}

\section{Introduction}
Personalized feedback is known to significantly enhance student achievement \cite{bloom-1984-the}.
However, providing real-time, individualized feedback at scale in traditional classroom settings is challenging due to limited resources.
Large language models (LLMs) can be particularly beneficial to address this challenge by enabling real-time feedback in educational settings \cite{yan-2024-practical, kasneci-2023-ChatGPT, wang-demszky-2023-chatgpt}.
However, LLMs often struggle to generate constructive feedback within educational contexts.
Unlike human feedback, which consistently identifies areas for improvement, LLM-generated feedback frequently fails to effectively highlight students' weaknesses \cite{behzad-etal-2024-assessing}. 
Therefore, it is essential to identify the advantages and limitations of LLMs as English writing tutors and to develop methods for providing effective feedback for students.

The evaluation of LLMs for educational purposes differs significantly from their general-purpose evaluation. 
General-purpose LLM evaluation primarily focuses on assessing the quality of responses \cite{wang-etal-2023-cue, zheng-2024-judging, chang-2024-survey}. 
However, as \citet{lee-2023-evaluating} emphasizes, merely evaluating the final output quality is insufficient to capture the full dynamics of human-LLM interactions. 
In particular, educational feedback needs a more nuanced consideration of factors beyond traditional metrics. 
It also requires the expertise of education professionals to evaluate the learning process and outcomes due to its inherent challenges. 
Our work incorporates metrics specifically tailored to pedagogical considerations, moving beyond traditional evaluation methods by involving real-world education stakeholders to better assess student-LLM interactions.

In summary, the main contributions of this work are as follows:
\begin{itemize}
    \item We explore the role of LLM as tutors in generating essay feedback.
    \item We introduce an educational evaluation metric customized for EFL writing education. 
    \item We assess student-LLM interactions by involving real-world educational stakeholders.
\end{itemize}


\section{LLMs as EFL Tutors: Early Insights}
In this section, we report preliminary findings that display both the advantages and limitations of LLM-as-a-tutor.  
\subsection{Advantage of LLM-as-a-tutor}
We conduct a group interview with six EFL learners and a written interview with three instructors to explore the needs for feedback generation. To reflect the perspectives of key stakeholders in EFL writing education, we recruit undergraduate EFL learners and instructors from a college EFL center. 
The use of LLM-as-a-tutor presents a significant opportunity for EFL learners by enabling real-time feedback at scale.
While all students expressed a strong need for both rubric-based scores and feedback, only two of them had previously received feedback from their instructors.
Students are particularly interested in receiving immediate scores and feedback, allowing them to identify weaknesses in their essays and refine them through an iterative process. 

\subsection{Limitation of LLM-as-a-tutor}
We conduct an experiment using \texttt{gpt-3.5-turbo} to generate essay feedback. 
The model is configured to act as an English writing teacher and provide feedback based on an EFL writing scoring rubric \cite{cumming-1990-expertise, ozfidan-2022-assessment}. 
Detailed experimental settings and prompts are described in Appendix \S\ref{sec:experimental_setting}. 
We ask 21 English education experts to evaluate the feedback on a 7-point Likert scale, focusing on tone and helpfulness.
The experts rate the feedback's positiveness at 5.93 and directness at 3.72. 
This result indicates \texttt{gpt-3.5-turbo}'s inherent tendency to generate positive feedback.
However, previous research and our qualitative interviews suggest that EFL learners prefer direct and negative feedback ~\cite{ellis-2009-typology, saragih-2021-written}.
Moreover, the experts found the feedback from \texttt{gpt-3.5-turbo} less helpful, with an average helpfulness rating of 3.41 out of 7.



\begin{figure*}[ht!]
    \centering
    \subfloat{\centering\includegraphics[width=0.45\textwidth]{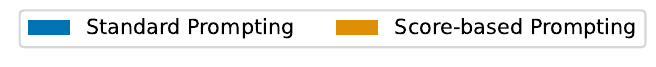}}
    \vspace{-5mm}
    \addtocounter{subfigure}{-1}
    \subfloat{\centering\includegraphics[width=0.15\textwidth, valign=t]{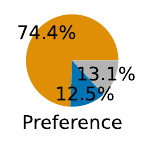}\hspace{7mm}\label{fig:feedback_preference}}
    \subfloat{\centering\includegraphics[width=0.165\textwidth, valign=t]{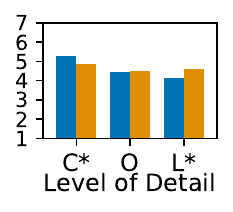}\hspace{5mm}\label{fig:feedback_evaluation_detail}}
    \subfloat{\centering\includegraphics[width=0.165\textwidth, valign=t]{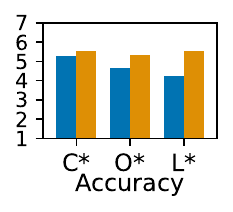}\hspace{6mm}\label{fig:feedback_evaluation_accuracy}}
    \subfloat{\centering\includegraphics[width=0.165\textwidth, valign=t]{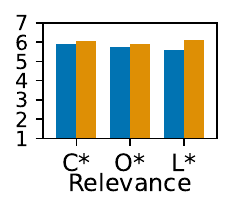}\hspace{6mm}\label{fig:feedback_evaluation_relevance}}
    \subfloat{\centering\includegraphics[width=0.165\textwidth, valign=t]{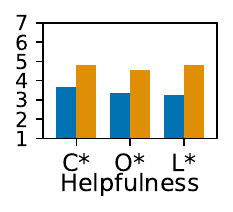}\label{fig:feedback_evaluation_helpfulness}}
    \vspace{-5mm}
    \subfloat{\centering\includegraphics[width=0.2\textwidth, valign=t]{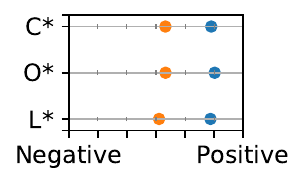}\label{fig:feedback_analysis_positive}}
    \subfloat{\centering\includegraphics[width=0.2\textwidth, valign=t]{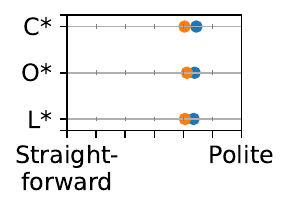}\label{fig:feedback_analysis_polite}}
    \subfloat{\centering\includegraphics[width=0.2\textwidth, valign=t]{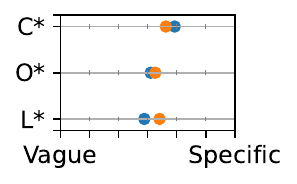}\label{fig:feedback_analysis_specific}}
    \subfloat{\includegraphics[width=0.2\textwidth, valign=t]{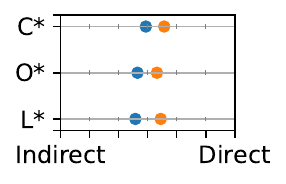}\label{fig:feedback_analysis_direct}}
    \subfloat{\includegraphics[width=0.2\textwidth, valign=t]{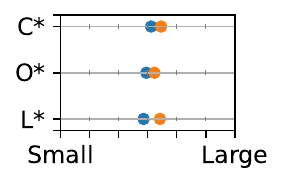}\label{fig:feedback_analysis_extensive}}
    \vspace{-4mm}
    \caption{Evaluation results on quality and characteristic of two rubric-based feedback with standard prompting and score-based prompting in a 7-point Likert scale. 
    C, O, and L denote Content, Organization, and Language, respectively. 
    Asterisk denotes statistical significance tested by the paired T-test at $p$ level of $< 0.05$.}
    \label{fig:feedback_evaluation_criteria}
\end{figure*}
\subsection{Mitigating Limitation}
To address the limitations of standard prompting in generating effective feedback for EFL learners, we propose a score-based prompting method that involves informing the model of a student's essay weakness using rubric-based scores.
While models like \texttt{gpt-3.5-turbo}, trained with reinforcement learning from human feedback, generally align with human preferences in broad contexts, they may not always provide the most constructive feedback for EFL learners who need more targeted guidance. 
These models tend to generate positive and indirect feedback, which, though satisfactory in general contexts, may not be as effective for learners who need more targeted and constructive guidance.
Therefore, we suggest score-based prompting method, leveraging rubric-based scores for LLM self-refinement of feedback generation \cite{pan-etal-2024-automatically}. 

Score-based prompting method uses predicted scores and rubric explanations to generate feedback on students' essays. 
Student's essays are scored on three rubrics (content, organization, and language) by the state-of-the-art automated essay scoring model \cite{yoo-2024-dress}.
We assume this scoring information can guide the model in generating feedback that is more aligned with students' needs. 
Detailed prompt is described in Appendix \S\ref{sec:experimental_setting}.

\section{Student-LLM Interaction Evaluation}
\subsection{Annotator Details}
\label{sec:annotator}
We explore student-LLM interaction of 33 EFL learners and gather evaluations from 21 English education experts, who are key stakeholders in EFL writing education. 
These experts hold Secondary School Teacher's Certificates (Grade II) for English, licensed by the Ministry of Education, Republic of Korea. 
The student cohort comprises 32 Korean students and one Italian student, with a gender distribution of 12 females and 21 males. 
While participating in EFL writing courses, students independently write their essays, which are then subjected to LLM-generated feedback. This feedback is produced by \texttt{gpt-3.5-turbo} using score-based prompting, and is delivered through the RECIPE~\cite{han-2023-recipe} platform as part of their coursework. 

\subsection{Evaluation Details}
\begin{table*}[htb!]
\centering
\begin{tabular}{@{}l|cc|l@{}}
\toprule
\textbf{Criteria} & \textbf{Target} & \textbf{Perspective} & \textbf{Metric} \\ \midrule
1. Quality         & Output  & Teacher & Level of detail, Accuracy, Relevance, Helpfulness \\ \midrule
2. Characteristic &
  Output &
  Teacher &
  \begin{tabular}[c]{@{}l@{}}Negative-Positive, Straightforward-Polite, \\ General-Specific, Indirect-Direct, Small-Large\end{tabular} \\ \midrule
3. Learning outcome & Process & Student & Essay quality improvement, Understanding          \\ \bottomrule
\end{tabular}%
\caption{Evaluation metrics constructed upon targets, perspectives, and criteria}
\label{tab:evaluation_metric}
\end{table*}
We introduce educational metrics specifically designed to assess student-LLM interactions within the context of EFL writing education (Table \ref{tab:evaluation_metric}). 
These metrics are constructed by adapting \citet{lee-2023-evaluating}'s framework to fit the EFL writing settings, focusing on targets (\S~\ref{sec:target}), perspectives (\S~\ref{sec:perspective}), and criteria (\S~\ref{sec:criteria}). 
\subsubsection{Targets}
\label{sec:target}
We identify two primary aspects for evaluating student-LLM interactions: \textit{output} and \textit{process}. \textit{Output} refers to the LLM's generated feedback that students receive, while \textit{process} encompasses the development of students' essays, their comprehension, and overall progress during the interaction.

\subsubsection{Perspectives}
\label{sec:perspective}
The evaluation involves the two main stakeholders in EFL education: \textit{students} and \textit{teachers}. While students may favor LLMs that provide immediate, correct answers, this approach may not be pedagogically optimal. Therefore, it is crucial to incorporate \textit{teachers'} perspectives when assessing the \textit{quality} and \textit{characteristics} of LLM-generated feedback.

\subsubsection{Criteria}
\label{sec:criteria}
We first evaluate student-LLM interactions using three key criteria: \textit{quality}, \textit{characteristics}, and \textit{learning outcomes}. 

\paragraph{Quality}
\label{sec:evaluation_criteria}
For quality assessment, we adapt evaluation criteria from LLM response assessments~\cite{zheng-2024-judging}, re-defining those criteria to suit our domain of feedback generation: level of detail, accuracy, relevance, and helpfulness. 

\begin{tcolorbox}[top=1pt, left=1pt, right=1pt, bottom=1pt]
\begin{itemize}[leftmargin=10pt, nosep]
    \item \textbf{Level of detail}: The feedback is specific, supported with details.
    \item \textbf{Accuracy}: The feedback content provides accurate information according to the essay.
    \item \textbf{Relevance}: The feedback is provided according to the understanding of the essay criteria.
    \item \textbf{Helpfulness}: The feedback is helpful for students to improve the quality of writing.
\end{itemize}
\end{tcolorbox}

\paragraph{Characteristics}
\label{sec:feedback_type}
Building on previous studies in education, we propose five characteristics to analyze the type of feedback in the context of English writing education. 
These criteria include: negative $\leftrightarrow$ positive~\cite{cheng-2022-teachers}, straightforward $\leftrightarrow$ polite~\cite{danescu-niculescu-mizil-etal-2013-computational, lysvaag-1975-verbs}, vague $\leftrightarrow$ specific~\cite{leibold-2015-art}, indirect $\leftrightarrow$ direct~\cite{eslami-2014-effects, van-2012-evidence}, small $\leftrightarrow$ large~\cite{liu-2015-methodological}. See Table \ref{tab:feedback_type_explanation} for more detailed explanations and examples.

\begin{tcolorbox}[top=1pt, left=1pt, right=1pt, bottom=1pt]
\begin{itemize}[leftmargin=10pt, nosep]
    \item \textbf{Negative $\leftrightarrow$ Positive}: Is the tone of feedback positive?
    \item \textbf{Straightforward $\leftrightarrow$ Polite}: Is the feedback polite?
    \item \textbf{Vague $\leftrightarrow$ Specific}: Is the feedback specific?
    \item \textbf{Indirect $\leftrightarrow$ Direct}: Is the feedback direct?
    \item \textbf{Small $\leftrightarrow$ Large}: How extensive is the quantity of feedback provided?
\end{itemize}
\end{tcolorbox}

\paragraph{Learning Outcome}
We assess the impact of student-LLM interaction on learning outcomes. 
Students assess their own learning progress by comparing their improvement before and after receiving feedback from the LLM.
After engaging with LLM-as-a-tutor to revise their essays, students reflect on their learning process through a questionnaire. The detailed questions are provided in Appendix \S\ref{tab:learning_outcome_question}.

\subsection{Results}
In this section, we report the results of standard and score-based prompting across three criteria: \textit{quality}, \textit{characteristic}, and \textit{learning outcome}.
\paragraph{Quality} 
Four figures in the top row in Figure \ref{fig:feedback_evaluation_criteria} present the quality evaluation results for the two types of feedback. 
Score-based prompting outperforms standard prompting in terms of accuracy, relevance, and helpfulness, achieving statistical significance across all rubrics.
Feedback generated by standard prompting varies in the level of detail (4.16 -- 5.28), while score-based prompting produces consistently detailed feedback (4.48 -- 4.86). Moreover, feedback from standard prompting tends to be overly detailed in summarizing the essay, which is not perceived as constructive (see examples in Table~\ref{tab:essaycot_example_quality_content}). Further qualitative analysis is described in Appendix \S\ref{tab:sample-level-efg-quality}.

\paragraph{Characteristic}
We evaluate feedback using five metrics tailored to English writing education.
Score-based prompting generates more negative, straightforward, direct, and extensive feedback compared to standard prompting across all rubrics (see the figures located in the lower section of Figure~\ref{fig:feedback_evaluation_criteria}).
Specifically, feedback from standard prompting tends to generate general compliments rather than constructive criticism.
In contrast, feedback from score-based prompting is notably more concise, delivering more content in significantly fewer tokens (70.46 vs. 79.19) and sentences (4.20 vs. 5.04). 
To further support the results, we also conduct a qualitative analysis of the feedback characteristics on Negative $\leftrightarrow$ Positive and Straightforward $\leftrightarrow$ Polite (Appendix~\S\ref{tab:sample-level-efg-type}). 

As a result, 74.38\% of teacher annotators prefer feedback from score-based prompting, compared to only 12.50\% who favor feedback from standard prompting (See pie chart in Figure \ref{fig:feedback_evaluation_criteria}). The remaining 13.12\% report no difference between the two feedback types. This preference is statistically significant at a $p$ level of < 0.05 using the Chi-squared test, with a fair agreement among annotators (Fleiss Kappa 0.22). 

\paragraph{Learning Outcome}
\begin{figure}[tb!]
    \centering
    \vspace{-2mm}
    \includegraphics[width=0.7\columnwidth]{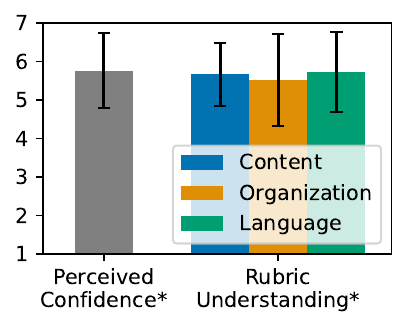}
    \vspace{-2mm}
    \caption{Learning outcome}
    \vspace{-5mm}
    \label{fig:student_response_thought}
\end{figure}
The feedback provided through score-based prompting leads to a significant improvement in students' confidence regarding the quality of their essays and their  understanding of each rubrics  (Figure~\ref{fig:student_response_thought}).
On average, EFL learners express high satisfaction with the LLM-generated feedback, rating 6.0 for  \textit{quality} and 6.03 for \textit{characteristics} on a 7.0 scale. 
These results are statistically significant, tested by the Wilcoxon test at $p$ value of < 0.05. 
Such strong positive response underscores the potential of score-based prompting on both student confidence and satisfaction, highlighting its potential as a valuable tool for enhancing writing instruction in EFL contexts.

\section{Conclusion}
This paper advances EFL writing education by generating and evaluating feedback tailored to students' needs, incorporating pedagogical principles, and involving real-world educational stakeholders.  
Our focus on essay feedback through LLM-as-a-tutor aims to more effectively support EFL students in their writing process.
In the future, we plan to customize the LLM-as-a-tutor to provide individualized support. 
For instance, our evaluation metric and dataset can be utilized to personalize feedback, aligning with students' varying preferences.
This customization would allow LLM-as-a-tutor to adapt to the specific needs and desires of each student, thereby enhancing the learning experience. 
In addition, we anticipate advancing the LLM-as-a-tutor by enabling real-time detection of each student's knowledge state.
Ultimately, we envision personalized LLM agents in EFL education, offering tailored support to each learner based on their unique needs.

\section*{Limitations}
We utilize ChatGPT, a black-box language model, for feedback generation.
This results in a lack of transparency in our system, as it does not provide explicit justifications or rationales for the generated feedback.
We acknowledge the importance of and the need for continued research aimed at developing models that produce more explainable feedback, thereby opening avenues for future exploration.

\section*{Ethics Statement}
We expect that this paper will make a significant contribution to the application of NLP for good, particularly in the domain of NLP-driven assistance in EFL writing education.
All studies are conducted with the approval of our institutional review board (IRB).
We ensured non-discrimination across all demographics, including gender and age.
We set the wage per session to be above the minimum wage in the Republic of Korea in 2023 (KRW 9,260 $\approx$ USD 7.25)\thinspace\footnote{\url{https://www.minimumwage.go.kr/}}.
Participation in the experiment was entirely voluntary, with assurance that their choice would not influence their academic scores or grades.


\bibliography{latex/anthology, latex/custom}

\newpage
\clearpage
\section*{Appendix}
\appendix
\section{Essay Feedback Generation Model}
\label{sec:experimental_setting}
The essay feedback generation experiments were conducted with \texttt{gpt-3.5-turbo} (0301 version) with Azure OpenAI API. 
To provide consistent feedback among students, we opted for a temperature setting of 0. This deterministic approach ensures that our system remains uniform, akin to evaluations from a single, consistent instructor.
Below is the prompt template we used for feedback generation. 
\begin{tcolorbox}[top=1pt, left=1pt, right=1pt, bottom=1pt]
\textbf{Standard Prompting} \\\\
    You are an English writing teacher;  \\give feedback on this argumentative essay with three rubrics: content, organization, and language.

    \texttt{\$\{rubric explanation\}} \\
    \texttt{\$\{essay prompt\}} \\
    \texttt{\$\{student's essay\}}
\end{tcolorbox}

\begin{tcolorbox}[top=1pt, left=1pt, right=1pt, bottom=1pt]
\textbf{Score-based Prompting} \\\\
    You are an English writing teacher; \\
    \colorbox{yellow!50}{according to the provided score}, give feedback on this argumentative essay with three rubrics: content, organization, and language.

    \texttt{\$\{rubric explanation\}} \\
    \texttt{\$\{essay prompt\}} \\
    \texttt{\$\{student's essay\}} \\

    \colorbox{yellow!50}{Score} \\
    {\texttt{\$\{rubric-based essay scores\}}}

\end{tcolorbox}

\begin{table}[htb!]
\begin{tabularx}{\columnwidth}{@{}l|X@{}}
    \toprule
    \textbf{Rubric} & \textbf{Description} \\ \midrule
    \textit{Content} & Paragraph is well-developed and relevant to the argument, supported with strong reasons and examples. \\ \midrule
    \textit{Organization} & The argument is very effectively structured and developed, making it easy for the reader to follow the ideas and understand how the writer is building the argument. Paragraphs use coherence devices effectively while focusing on a single main idea.     \\ \midrule
    \textit{Language} & The writing displays sophisticated control of a wide range of vocabulary and collocations. The essay follows grammar and usage rules throughout the paper. Spelling and punctuation are correct throughout the paper.  \\ \bottomrule
\end{tabularx}
\caption{Rubric explanations}
\label{tab:rubric_explanation}
\end{table}

\section{Essay Feedback Evaluation Details}
\begin{table*}[htb!]
\small
\centering
\begin{tabularx}{\textwidth}{@{}l|X|X@{}}
\toprule
\textbf{Type}            & \textbf{Explanation}                                                                                                                                                      & \textbf{Example}                                                                                                                                                                                                                                                                                                                                                                                                                                                                                                                                                                                                                                                                                                    \\ \midrule \midrule
Negative        & Teachers’ comments indicate that there are some errors, problems, or weaknesses in students’ writing.                                                            & The essay lacks depth and development in its content.                                                                                                                                                                                                                                                                                                                                                                                                                                                                                                                                                                                                                                                      \\ \midrule
Positive        & The former refers to comments affirming that students’ writing has met a standard such as ``good grammar'', ``clear organization'', and ``the task is well achieved''. & The essay is very well-organized and effectively structured.                                                                                                                                                                                                                                                                                                                                                                                                                                                                                                                                                                                                                                               \\ \midrule \midrule
Polite          & Politeness includes hedge expressions, modal verbs, positive lexicon, and 1st person pronouns.                                                                   & However, the essay could benefit from more elaboration and development of each point.                                                                                                                                                                                                                                                                                                                                                                                                                                                                                                                                                                                                                      \\ \midrule
Straightforward & Straightforward includes factuality expression and negative lexicon                                                                                              & The essay lacks depth and analysis.                                                                                                                                                                                                                                                                                                                                                                                                                                                                                                                                                                                                                                                                        \\ \midrule \midrule
Vague           & Feedback is vague in its suggestions for ways a student can enhance their work.                                                                                  & There are some grammar errors.                                                                                                                                                                                                                                                                                                                                                                                                                                                                                                                                                                                                                                                                             \\ \midrule
Specific        & Feedback is specific in its suggestions for ways a student can enhance their work.                                                                               & There are some split infinitives in the paper. Check out more information about split infinitives in the courseroom folder titled Writing Resources.                                                                                                                                                                                                                                                                                                                                                                                                                                                                                                                                                       \\ \midrule \midrule
Indirect        & The teacher indicates in some way that an error exists but does not provide the correction, thus leaving it to the student to find it.                           & However, the essay could benefit from more examples and evidence to further strengthen the argument \ldots                                                                                                                                                                                                                                                                                                                                                                                                                                                                                                                                                                                                   \\ \midrule
Direct          & The teacher provides the student with the correct form.                                                                                                          & In the third paragraph, the phrase `unsatisfied things' could be more specific and descriptive.                                                                                                                                                                                                                                                                                                                                                                                                                                                                                                                                                                                                            \\ \midrule \midrule
Small           & Feedback with a small quantity contains less content.                                                                                                            & The essay provides a clear argument and supports it with well-developed paragraphs that are relevant to the topic. The reasons and examples provided are strong and effectively demonstrate the writer's opinion. The essay effectively addresses the prompt and provides a well-rounded argument.                                                                                                                                                                                                                                                                                                                                                                                                         \\ \midrule
Large           & Feedback with a large quantity contains more extensive content in the feedback.                                                                                  & The essay provides a clear and well-supported argument on the topic of whether young children should spend most of their time playing or studying. The writer presents two strong reasons for their opinion that playing is better for young children. The first reason is that playing is a way of studying, as it helps children learn how to communicate and collaborate with others. The second reason is that young children are not yet mature enough for formal education, and forcing them to learn before they are ready can lead to a decline in their interest in learning. The writer supports their argument with specific examples and uses clear and concise language throughout the essay. \\ \bottomrule
\end{tabularx}
\caption{Explanation and example of feedback types}
\label{tab:feedback_type_explanation}
\vspace{-2mm}
\end{table*}

\subsection{Sample-level Analysis on Essay Feedback Evaluation}
\label{tab:sample-level-efg}
\subsubsection{Quality}
\label{tab:sample-level-efg-quality}
Table \ref{tab:essaycot_example_quality_language} shows two different language feedback examples for the same essay with a score of 2.5 out of 5.0. These examples are generated using different prompts: a standard prompt and a score-based prompt.
The green text indicates detailed support and examples provided by the essay (level of detail), and the blue text describes the overall evaluation of the essay regarding the language criterion. By comparing blue text, score-based prompting suggests the improvements (helpfulness) such as \textit{`errors and awkward phrasing'} and \textit{`punctuation and capitalization'}, while standard prompting only praises language use such as \textit{`vocabulary and collocations'}. 
Considering that the language score of the essay is 2.5 out of 5.0, the feedback generated by score-based prompting appears to be more accurate.
The orange text in the feedback generated by the standard prompt is irrelevant to the language criterion (relevance) and has similar expressions to an organization explanation in Table~\ref{tab:rubric_explanation}.
We assume that score-based prompting, providing more detailed, relevant, and accurate feedback, will be more helpful to EFL students in improving their essays.
This again highlights the necessity of the score-based prompting process.

The green text in Table \ref{tab:essaycot_example_quality_content} represents detailed support and examples, while the pink text suggests improvements for the essay, which align with level of detail and helpfulness criteria, respectively. Feedback generated with standard prompting describes specific details of the content of the essay. However, it only summarizes the content of the essay and provides praise, instead of suggestions or criticism. On the other hand, feedback with score-based prompting points out that \textit{`the essay lacks depth and analysis'} and \textit{`could benefit from expanding on their ideas and providing more specific examples'}.
\subsubsection{Characteristics}
\label{tab:sample-level-efg-type}
Table \ref{tab:essaycot_example_characteristic} 
presents two different content feedback examples for the same essay, scored 2.0 out of 5.0. Positive feedback is highlighted in blue text, while negative feedback is in orange. Negative feedback identifies errors in the essay, helping students recognize and address their weaknesses. Positive feedback, on the other hand, affirms that students' writing meets certain standards through praise~\cite{cheng-2022-teachers, hyland-2001-sugaring}. When comparing feedback generated by the two prompting methods, score-based prompting focuses more on negative feedback, while standard prompting only emphasizes positive aspects. Since overuse of negative feedback may discourage students' self-efficacy, teachers often precede negative feedback with positive feedback or tone down the negative feedback~\cite{hyland-2001-sugaring} with hedge expressions such as `could', which is written as green text in Table~\ref{tab:essaycot_example_characteristic}.

\begin{table*}[ht!]
\small
\begin{tabularx}{\textwidth}{@{}X|X@{}}
\toprule
\multicolumn{1}{c|}{\textbf{Standard Prompting}}                                                                                                                                                                                                                                                                                                                                                                                                                                                    & \multicolumn{1}{c}{\textbf{Score-based Prompting}}                                                                                                                                                                                                                                                                                                                                                                                                                                                                                                                                                        \\ \midrule
The content of the essay is well-developed and relevant to the argument. The writer has identified \textcolor[HTML]{029e73}{three major problems in their country} and has provided specific reasons and details to support their opinion. The writer has effectively highlighted \textcolor[HTML]{029e73}{the gender problem, inequality problem, and offensive replying culture in some communities}. The writer has also provided examples of how these problems are \textcolor[HTML]{029e73}{threatening their society and splitting people up}. Overall, the content of the essay is strong and persuasive. & The essay has some relevant points, \textcolor[HTML]{0173b2}{but the content is not well-developed}. The writer mentions \textcolor[HTML]{029e73}{three problems in Korea}, \textcolor[HTML]{0173b2}{but only briefly touches on each one without providing strong reasons or examples} to support their argument. The essay \textcolor[HTML]{0173b2}{lacks depth and analysis}, and the writer \textcolor[HTML]{0173b2}{could benefit from expanding on their ideas and providing more specific details}. \\ \bottomrule
\end{tabularx}
\caption{Example of quality evaluation on content feedback. The feedback is generated based on the student's essay, whose content score is 2.5 out of 5.0. The colors of the texts stand for quality evaluation criteria as follows: green (level of detail), and blue (helpfulness).}
\label{tab:essaycot_example_quality_content}
\end{table*}

\begin{table*}[ht!]
\small
\centering
\begin{tabularx}{\textwidth}{@{}X|X@{}}
\toprule
\multicolumn{1}{c|}{\textbf{Standard Prompting}}                                                                                                                                                                                                                                                                                                                                                                                                                                                    & \multicolumn{1}{c}{\textbf{Score-based Prompting}}                                                                                                                                                                                                                                                        \\ \midrule
 The language used in the essay is \textcolor[HTML]{0173b2}{sophisticated} and displays \textcolor[HTML]{0173b2}{a wide range of vocabulary and collocations}. The writer follows grammar and usage rules throughout the paper, and \textcolor[HTML]{0173b2}{spelling and punctuation are correct}. The essay is well-written and \textcolor[HTML]{de8f05}{easy to read}. Overall, the language used in the essay is strong and \textcolor[HTML]{de8f05}{effectively communicates the writer's argument}.

 & The essay displays some control of vocabulary and grammar, but there are also \textcolor[HTML]{0173b2}{several errors and awkward phrasings} throughout. For example, the writer uses the phrase (\textcolor[HTML]{029e73}{`nowdays' instead of `nowadays'}), and there are several instances of awkward phrasing and word choice. Additionally, there are \textcolor[HTML]{0173b2}{several errors in punctuation and capitalization}. The essay would benefit from more careful proofreading and editing, as well as more attention to precise and effective language use. \\ \bottomrule
\end{tabularx}
\caption{Example of quality evaluation on language feedback. The feedback is generated based on the student's essay, whose language score is 2.5 out of 5.0. The colors of the texts stand for quality evaluation criteria as follows: green (level of detail), blue (helpfulness), and orange (relevance).}
\vspace{-1mm}
\label{tab:essaycot_example_quality_language}
\end{table*}
\begin{table*}[ht!]
\small
\centering
\begin{tabularx}{\textwidth}{@{}X|X@{}}
\toprule
\multicolumn{1}{c|}{\textbf{Standard Prompting}}                                                                                                                                                                                                                                                                                                                                                                                                                                                    & \multicolumn{1}{c}{\textbf{Score-based Prompting}}                                                                                                                                                                                                                                                        \\ \midrule
The essay  \textcolor[HTML]{0173b2}{effectively addresses the prompt} by discussing three qualities of a good friend: honesty, sharing, and competition. Each quality is  \textcolor[HTML]{0173b2}{well-developed and supported} with specific reasons and examples. The writer's argument is  \textcolor[HTML]{0173b2}{clear and persuasive}, making a strong case for why these qualities are important in a friendship. Overall, the content is  \textcolor[HTML]{0173b2}{well-organized and relevant} to the topic at hand.
 & The essay provides \textcolor[HTML]{0173b2}{a clear and concise argument} about the qualities of a good friend. The writer has identified three key qualities - honesty, sharing, and competition - and has provided \textcolor[HTML]{0173b2}{specific reasons and examples} to support each point. \textcolor[HTML]{de8f05}{However, the essay} \textcolor[HTML]{029e73}{could} \textcolor[HTML]{de8f05}{benefit from more elaboration and development of each point.} For example, the writer \textcolor[HTML]{029e73}{could} \textcolor[HTML]{de8f05}{provide more specific examples} of how honesty is important in a friendship, or how competition can lead to personal growth. Additionally, the essay could \textcolor[HTML]{de8f05}{benefit from a stronger conclusion that summarizes the main points and provides a final thought on the topic.}\\ \bottomrule
\end{tabularx}
\caption{Example of feedback type analysis on language feedback. The feedback is generated based on the student's essay, whose language score is 2.0 out of 5.0. The colors of the texts stand for feedback type as follows: blue (positive), green (polite), and orange (negative).}

\vspace{-1mm}
\label{tab:essaycot_example_characteristic}
\end{table*}

\section{Questionaire for Learning Outcome}
\label{tab:learning_outcome_question}
    Please answer these questions AFTER finishing the main exercise.
    \begin{enumerate}
        \item My confidence in the quality of the essay increased after the exercise.
        \item My understanding of the content criteria increased after the exercise.
        \item My understanding of the organization criteria increased after the exercise.
        \item My understanding of the language criteria increased after the exercise.
        \item Please rate the appropriateness of the style or tone of the AI-based feedback.
        \item Please rate the overall performance of AI-based scoring.
        \item Please rate the overall quality of AI-based feedback.
        \item Please freely share your thoughts regarding the exercise.
    \end{enumerate}

\end{document}